\documentclass[]{ceurart}
\usepackage{graphics}
\usepackage{xcolor}

\begin{document}

\copyrightyear{2021}
\copyrightclause{Copyright for this paper by its authors.
  Use permitted under Creative Commons License Attribution 4.0
  International (CC BY 4.0).}

\conference{De-Factify: Workshop on Multimodal Fact Checking and Hate Speech Detection, co-located with AAAI 2022. 2022
Vancouver, Canada}

\title{BLUE at Memotion 2.0 2022: You have my Image, my Text and my Transformer}

\author[1,2]{Ana-Maria Bucur}[
email=ana-maria.bucur@drd.unibuc.ro,
]
\address[1]{Interdisciplinary School of Doctoral Studies, University of Bucharest, Romania}
\address[2]{Universitat Politècnica de València, Spain}

\author[3]{Adrian Cosma}[
email=cosma.i.adrian@gmail.com,
]
\address[4]{Faculty of Automatics and Control, University Politehnica of Bucharest, Romania}

\author[4]{Ioan-Bogdan Iordache}[
email=iordache.bogdan1998@gmail.com,
]
\address[3]{University of Bucharest, Romania}

\begin{abstract}
    Memes are prevalent on the internet and continue to grow and evolve alongside our culture. An automatic understanding of memes propagating on the internet can shed light on the general sentiment and cultural attitudes of people.
    In this work, we present team BLUE's solution for the second edition of the MEMOTION shared task. We showcase two approaches for meme classification (i.e. sentiment, humour, offensive, sarcasm and motivation levels) using a text-only method using BERT, and a Multi-Modal-Multi-Task transformer network that operates on both the meme image and its caption to output the final scores. In both approaches, we leverage state-of-the-art pretrained models for text (BERT, Sentence Transformer) and image processing (EfficientNetV4, CLIP). Through our efforts, we obtain first place in task A, second place in task B and third place in task C. In addition, our team obtained the highest average score for all three tasks.
\end{abstract}


\maketitle

\section{Introduction}
The concept of a meme was first introduced by Richard Dawkins \cite{r.dawkins1976the-selfish-gen}, as a fundamental unit of propagation of ideas and cultural information, similar to genes for transmitting genetic information across time. Dawkins proposed a memetic theory, in which memes have very similar patterns evolution by natural selection as genes, in which memes evolve and replicate across time, through mutation and cross-over with other memes. While this theory was criticized \cite{solitary-self} from the onset, it remained a valuable tool for viral marketing, social analytics and understanding of cultural evolution across history. 

A widespread usage of the term "meme", aside from the more academic definitions from memetic theories, is that of internet memes in the sense of catch-phrases, images, gifs and videos. Internet memes "evolved" from simple images templates into modern, ironic and absurdist images, similar to tendencies in postmodern art (i.e. "deep fried" memes, dank memes). Websites such as 9gag, Tumblr, and Reddit were at the forefront of internet meme propagation and mainstream spread. 

Internet memes continue to grow and evolve alongside our culture. An automatic understanding of memes propagating on the internet can shed light on people's general sentiments and cultural attitudes. 

In this work, we present team BLUE's solution to the 2022 edition of the MEMOTION 2.0 shared task \cite{memotiontask}. We focused our efforts on two main approaches: i) text-based fine-tuning using BERT and ii) a Multi-Modal-Multi-Task transformer that uses features from both images and text. Furthermore, we provide ablation studies on different modalities and training parameters, and show that there is not one combination of modalities suitable for all tasks.

In the following sections, we make an overview of related methods from the previous shared task, briefly describe the task, describe the available training, validation and testing data for the current edition of the shared task. Finally, we describe our two approaches and report our results on both validation and test sets. 

\section{Related Work}
With the increase in popularity of social media websites (e.g. Facebook, Reddit, Twitter), NLP researchers started using the textual data collected from these platforms for detecting emotions \cite{demszky-etal-2020-goemotions, jianqiang2018deep, alvarez-gonzalez-etal-2021-uncovering-limits}, offensive content \cite{davidson2017automated, zampieri-etal-2019-predicting, rosenthal-etal-2021-solid}, hate speech \cite{cao2020deephate, aluru2020deep}, humour \cite{REYES20121, weller-seppi-2019-humor, ortega2018uo}, sarcasm \cite{plepi-flek-2021-perceived-intended-sarcasm, hazarika-etal-2018-cascade, bamman2015contextualized}, pejorative language \cite{dinu-etal-2021-computational-exploration}, inspirational content \cite{ignat2021detecting}, optimism \cite{caragea-etal-2018-exploring, ruan-etal-2016-finding} and the manifestations of mental health problems such as depression \cite{husseini-orabi-etal-2018-deep, bucur2021early}, suicide ideation \cite{sawhney-etal-2021-suicide, coppersmith2015quantifying} and anxiety \cite{shen-rudzicz-2017-detecting}. Researchers explored the online content from social media even further and began focusing on the multi-modal data \cite{10.1145/2964284.2964321, gomez2020exploring}, including internet memes. Efforts to automatically detect the offensive \cite{suryawanshi-etal-2020-multimodal} or harmful memes \cite{pramanick-etal-2021-momenta-multimodal} are being made to help the content moderators in charge of removing the posts containing hate speech. 

Several competitions took advantage of the high availability of internet memes and used the multi-modal data for various tasks:
DANKMEMES from Evalita 2020 \cite{Miliani2020DANKMEMESE}, MEMOTION from SemEval 2020 \cite{sharma-etal-2020-semeval}, The Hateful Meme Challenge \cite{kiela2020hateful}, Fine Grained Hateful Memes Detection Shared Task from The 5th Workshop on Online Abuse and Harms \cite{mathias-etal-2021-findings}, Detection of Persuasion Techniques in Texts and Images from SemEval 2021 \cite{dimitrov-etal-2021-semeval}, Multimodal Fact-Checking Task from the First Workshop on Multimodal Fact-Checking and Hate Speech Detection (De-Factify) \footnote{https://aiisc.ai/defactify/} and Multimedia Automatic Misogyny Identification (MAMI) from SemEval 2022 \footnote{https://github.com/MIND-Lab/MAMI}.

In the first iteration of the MEMOTION task at SemEval 2020 \cite{sharma-etal-2020-semeval}, the participating teams surpassed the baseline models by only a small percentage. Two participating teams used only the textual data extracted from the memes for all their experiments. The rest of the teams experimented with systems using visual-only or textual-only information or the fusion of both features. The majority of the participating teams relied on approaches based on pretrained models such as ResNet \cite{he2016deep}, VGG-16 \cite{DBLP:journals/corr/SimonyanZ14a} and Inception-ResNet \cite{szegedy2017inception} to extract the visual features. For textual information the teams used approaches based on Recurrent Neural Network architectures or pretrained transformer models such as BERT \cite{devlin-etal-2019-bert}. For task A, the system with the best performance used only the textual data from the internet memes for sentiment detection. For tasks B and C, the best performing systems used both image and text data for identifying the emotion of the memes.

In our participation in the MEMOTION 2.0 shared task, we used two approaches: a text-only approach using BERT and a Multi-Modal-Multi-Output transformer using both visual and textual features. For the second approach, we also used features extracted from CLIP, this model being suitable for multi-modal inputs.

\section{Task Description}
The second iteration of the MEMOTION shared task, previously conducted at SemEval 2020 \cite{sharma-etal-2020-semeval}, is comprised of three tasks for detecting the sentiment and the emotions of memes as described below:

\begin{itemize}
    \item \textbf{Task A: Sentiment Analysis}: Identify if a meme is positive, negative or neutral.
    \item \textbf{Task B: Emotion Classification}: Identify the emotion expressed by a meme: humour, sarcasm, offensive and motivation. A meme can convey more than one emotion.
    \item \textbf{Task C: Scales/Intensity of Emotion Classes}: Quantify to which extent a particular emotion is being expressed in a meme. The intensities are on a scale from 0 to 3 for humour, sarcasm and offensiveness (e.g. 0 - not funny, 1 - funny, 2 - very funny, 3 - hilarious) and only 0 and 1 for motivation (0 - not motivational, 1 - motivational).
\end{itemize}

The tasks are challenging, as identifying the sentiment and emotion in a meme is more complex than performing the same task only on textual data. For memes, comprised of image and text information, a multi-modal approach for understanding both visual and textual cues is needed. The dataset from the shared task contains memes with overlapping emotions, increasing the difficulty of the tasks. Most funny memes are also sarcastic, and some motivational memes are also offensive \cite{sharma-etal-2020-semeval}.

The teams' performance is evaluated by the weighted F1 score for task A. For tasks B and C, the weighted F1 score is computed for each subtask (humour, sarcasm, offensive, motivation), and the average F1 score of these subtasks is used to rank the systems.

\section{Data}

The dataset used in the MEMOTION 2.0 shared task \cite{memotiondata} is comprised of internet memes collected from the public domain. These memes were annotated by Amazon Mechanical Turk workers for sentiment and emotion. The dataset contains the image of the memes and the corresponding OCR extracted text. The annotators were also asked to provide the corrected text if the OCR extracted text was inaccurate.

The training set is comprised of 7K memes, and the validation and test splits contain 1.5K memes each. The distribution of labels for the three splits is presented in Table \ref{table:dataset}, the dataset is heavily imbalanced. For the sentiment labels, there are more positive memes than negative ones in the training and validation splits, while in the test split, there is only a very small number of memes with positive sentiment.

\begin{table}[!ht]
  \centering
  \resizebox{\linewidth}{!}{
  \begin{tabular}{lcccccccccccc}
    \toprule
     &  \multicolumn{4}{c}{Train data} & \multicolumn{4}{c}{Validation data} & \multicolumn{4}{c}{Test data} \\
    \midrule
     Label &  Negative  & Neutral  & Positive & &  Negative  & Neutral  & Positive &  & Negative  & Neutral  & Positive &    \\
    Sentiment       & 973 & 4510 & 1517 &  & 200 & 975 & 325 &  & 451 & 971 & 78 &  \\
    \midrule
    Label  &  0 & 1  & 2 & 3 &  0  & 1  & 2 & 3 & 0 & 1  & 2 & 3 \\
    Humour       & 918 & 3666 & 1865 & 551 & 229 & 745 & 419 & 107 & 62 & 892 & 398 & 148 \\
    Sarcasm    & 3871 & 1759 & 1069 & 301 & 804 & 388 & 246 & 62 & 185 & 248 & 892 & 175 \\
    Offensiveness    & 5182 & 1107 & 529 & 182 & 1110 & 238 & 107 & 45 & 943 & 457 & 87 & 13  \\
    Motivation    & 6714 & 286 & - & - & 1430 & 70 & - & - & 1480 & 20 & - & - \\
    \bottomrule
  \end{tabular}
  }
  \caption{
    The distribution of labels for sentiment and emotion in the MEMOTION 2.0 tasks. The dataset is heavily imbalanced.
  }
  \label{table:dataset}
\end{table}

\begin{figure}[hbt!]
    \centering
    \includegraphics[width=0.65\textwidth]{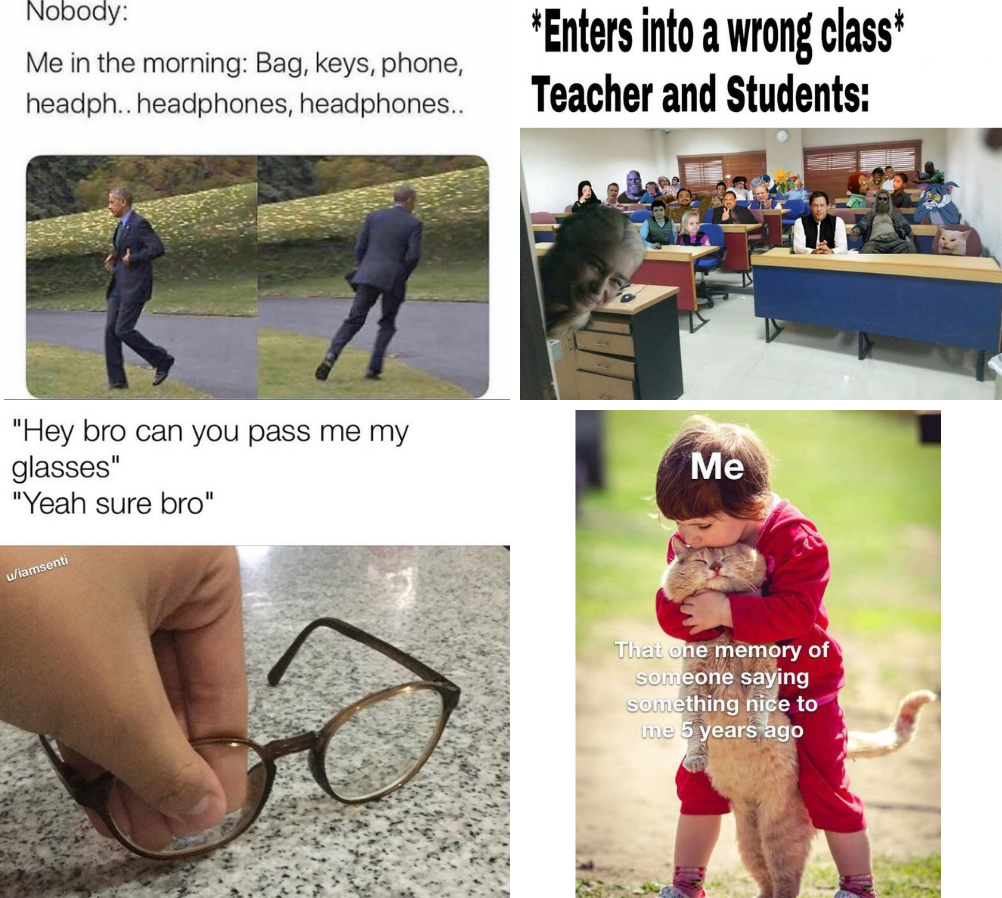}
    \caption{Example of memes from the dataset used in the MEMOTION 2.0 shared task. As seen in these examples, the memes cannot be understood without considering both the visual and textual cues.}
    \label{fig:example-memes}
\end{figure}

Given the multi-modal content found in internet memes, the textual or visual information alone may not be sufficient for identifying the sentiment and the emotions in a meme. Some examples from the dataset are presented in Figure \ref{fig:example-memes}, in which visual content is necessary for a complete understanding of the meme context. As such, we propose, alongside a text-only based method, a fusion approach combining both visual and textual information through a Multi-Modal-Multi-Task transformer.

\vspace{-6mm}

\section{Method}
In this section, we describe two methods for meme classification, as previously mentioned. We participated in all three tasks from the MEMOTION 2.0 shared task and used two approaches: a text-only meme classification using BERT \cite{DBLP:journals/corr/abs-1810-04805} and a Multi-Modal-Multi-Task transformer network.

\subsection{Text-Only Multi-Task Meme Classification}

Previous methods have shown that combining image features with textual features can bring more noise to the dataset \cite{bonheme-grzes-2020-sesam}, and for some tasks, the best results were achieved by text-only approaches \cite{keswani2020iitk}.

We propose a text-only multi-task method in which we extract textual features using a pretrained BERT model \cite{devlin-etal-2019-bert}. Neural network architectures based on transformers \cite{vaswani2017attention}, such as BERT, were shown to obtain great performance on many different Natural Language Processing tasks. Moreover, these results can be obtained by training such architectures on a large set of texts and then fine-tuning the model for different downstream tasks.

\textbf{Processing Pipeline.} Our implementation is based on a BERT base model provided by the HuggingFace library \cite{wolf2019huggingface}, consisting of a
$12$-layer transformer, with a hidden size of $768$ and $12$ attention heads. The OCR-extracted text of a meme is firstly tokenized using a SentencePiece \cite{kudo2018sentencepiece} tokenizer provided with the BERT model, then the list of tokens is passed through the encoder. The features extracted by BERT (the embedding of the \texttt{[CLS]} token) are passed through $5$ classification heads, corresponding to each of the emotions defined for a meme.
A classification head is implemented as a feed-forward layer with the output size specified by the task.

\textbf{Training details.} During training, we applied dropout with a rate of $0.1$ between the feature extractor and the classification heads. We used cross-entropy loss to compute the loss for each task. The training was done in mini-batches of size $16$ (each batch
containing examples for a single task), for $5$ epochs, saving the model with the best performance on the validation set. We fine-tuned the architecture end-to-end using the AdamW optimizer \cite{kingma2014adam}, with a learning rate of 0.00002, decreased at each step using
a linear scheduler, and no weight decay. For each task, we randomly oversampled instances from the training set, due to heavy dataset imbalance.

\subsection{Multi-Modal-Multi-Task Transformer (MMMT)}

\begin{figure}[hbt!]
    \centering
    \includegraphics[width=\textwidth]{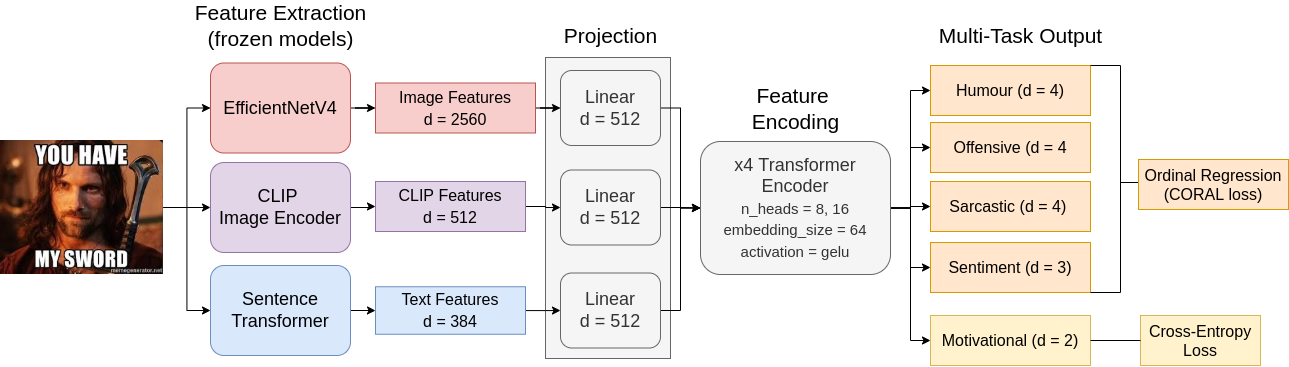}
    \caption{The general architecture of our Multi-Modal-Multi-Task model. We use image features from a pretrained EfficientNetV4 \cite{DBLP:journals/corr/abs-1905-11946}, CLIP \cite{DBLP:journals/corr/abs-2103-00020} features and sentence features from a pretrained sentence transformer \cite{DBLP:journals/corr/abs-1908-10084}. The features are passed through a 4-layer transformer encoder which outputs a classification head for each required aspect of the memes.}
    \label{fig:method}
\end{figure}

Internet memes are inherently multi-modal, often having a pop-culture image reference and a caption that accompanies the image, overlaid on top of it. While the meaning can be estimated using only the caption, images offer important additional context for higher-level semantic understanding and differentiating between types of memes (i.e. dank, deep-fried, "classical"), each with its predominant sentiment.

Several previous methods have reported using multi-modal approaches in the computational pipeline \cite{10.1145/2964284.2964321, gomez2020exploring,suryawanshi-etal-2020-multimodal,pramanick-etal-2021-momenta-multimodal}. In our pipeline, however, we explore semantic image features in two ways: i) direct image features provided by a pretrained EfficientNetV4 \cite{DBLP:journals/corr/abs-1905-11946} on ImageNet dataset \cite{deng2009imagenet}, and ii) features from the image encoder of CLIP \cite{DBLP:journals/corr/abs-2103-00020}. CLIP was trained on a large-scale multi-modal dataset of image-text pairs in a contrastive learning fashion for use in zero-shot image classification. Features extracted from CLIP have been shown to respond to multi-modal inputs \cite{goh2021multimodal}, such as the same concept being represented explicitly as an image, a high level representation of it (i.e. a sketch), or in written form. This makes CLIP suitable for use in our scenario. However, fine-tuning classification results are still lagging behind pretrained models trained only on images, so we also opted for using direct image features from EfficientNetV4 \cite{DBLP:journals/corr/abs-1905-11946}. 

\textbf{Processing Pipeline.} Our pipeline is described in Figure \ref{fig:method}. The first step in the computation is feature extraction using pretrained models. These models are frozen, and are not fine-tuned during training. We employ EfficientNetV4 \cite{DBLP:journals/corr/abs-1905-11946} and CLIP image encoder \cite{DBLP:journals/corr/abs-2103-00020} for image features, and a Sentence Transformer \cite{DBLP:journals/corr/abs-1908-10084} for processing the meme caption. Each of these models outputs a vector of different dimensionality, so we employ a different linear projection layer to change the dimensionality to a common 512-element vector. These 3 vectors are considered a set of features and are processed using a popular transformer architecture. The order of the different features does not matter in the final computation, and for that reason, we do not employ positional embeddings. The final features are averaged and are then followed by an output layer corresponding to each aspect of memes (humour, sarcasm, offensive, sentiment and motivation). For humour, sarcastic, offensive and sentiment heads, we employed CORAL \cite{coral2020} loss for ordinal regression, to consistently estimate the degree of humour, sarcasm, etc. 

\textbf{Training details.} The transformer network has 4 layers, with 8, 8, 16 and 16 attention heads, respectively. The internal embedding size is 64, and we used GELU activation \cite{DBLP:journals/corr/HendrycksG16}. We trained the network for 100 epochs, or until it overfits the validation set, with a batch size of 256 and Adam optimizer \cite{kingma2014adam}.  We employed a cyclical, triangular learning rate schedule \cite{DBLP:journals/corr/Smith15a}, with a step size of 5 epochs for a 10x increase in learning rate, and an initial learning rate of 0.0001. Since the training data is severely imbalanced, we oversampled minority classes.

\section{Experiments \& Results}
\newcommand{\resultsbyemotiontable}{
{\renewcommand{\arraystretch}{1.5}
\begin{table}[!ht]
  \centering
  \begin{tabular}{l|c|cc}
    \toprule
    Emotion             & Task &  Only Text & MMMT \\
    \midrule \midrule
    Sentiment & Task A & $0.5072$   & $\pmb{0.5318}$ \\
    \hline
    Humour & Task B     & $\pmb{0.9239}$   & $0.8111$ \\
    Humour & Task C     & $\pmb{0.4131}$   & $0.4036$ \\
    \hline
    Sarcasm & Task B    & $0.6386$   & $\pmb{0.8191}$ \\
    Sarcasm & Task C    & $0.1604$   & $\pmb{0.3083}$ \\
    \hline
    Offensive & Task B  & $\pmb{0.5581}$   & $0.485$ \\
    Offensive & Task C  & $\pmb{0.5045}$   & $0.485$ \\
    \hline
    Motivation & Tasks B \& C & $0.9764$   & $\pmb{0.98}$ \\
    \bottomrule
  \end{tabular}
  \caption{
    Weighted F1 scores computed for each emotion defined in the test dataset. 
  }
  \label{table:results-by-emotion}
\end{table}}}

\newcommand{\resultsbytasktable}{
{\renewcommand{\arraystretch}{1.5}
\begin{table}[!ht]
  \centering
  \begin{tabular}{l|cccc}
    \toprule
    Model           &  Task A    & Task B   & Task C   & Mean \\
    \midrule
    Only Text       & $0.5072$   & $\pmb{0.7743}$ & $0.5136$ & $0.5984$ \\
    MMMT    & $\pmb{0.5318}$   & $0.7738$ & $\pmb{0.5443}$ & $\pmb{0.6166}$ \\
    \bottomrule
  \end{tabular}
  \caption{
    Weighted F1 scores computed for each task for the test dataset.
  }
  \label{table:results-by-task}
\end{table}}}

\newcommand{\textonlymultivssingletable}{
{\renewcommand{\arraystretch}{1.5}
\begin{table}[!ht]
    \centering
    \begin{tabular}{l|ccccc}
        \toprule
        Setting     & Sentiment & Humour    & Sarcasm   & Offensive & Motivation \\
        \midrule
        Single-Task & $\pmb{0.5184}$  & $0.3682$  & $0.3641$  & $0.6145$  & $0.9285$ \\
        Multi-Task  & $0.5054$  & $\pmb{0.3752}$  & $\pmb{0.3973}$  & $\pmb{0.6148}$  & $\pmb{0.9286}$ \\
        \bottomrule
    \end{tabular}
    \caption{Comparing the weighted F1 scores computed on the validation dataset for tasks A and C,
    using the text-only approach and training the model either in the multi-task setting, or independently
    for each of the emotions}
    \label{table:text-only-single-vs-multi}
\end{table}}}

\newcommand{\textonlybinaryvsmulticlasstable}{
{\renewcommand{\arraystretch}{1.5}
\begin{table}[!ht]
    \centering
    \begin{tabular}{l|l|ccc}
         \toprule
         Target Task    & Setting       & Humour    & Sarcasm   & Offensive \\
         \midrule
         Task B         & single-task   & $\pmb{0.7817}$  & $\pmb{0.52}$    & \pmb{$0.6529$} \\
         Task B         & multi-task    & $0.6551$  & $0.5081$  & $0.5404$ \\
         Task C         & single-task   & $0.7688$  & $0.4746$  & $0.6316$ \\
         Task C         & multi-task    & $0.7676$  & $0.49$    & $0.6368$ \\
         \bottomrule
    \end{tabular}
    \caption{Performance of text-only models on predicting binary labels for humour, sarcasm and offensive emotions (task B). We compare the performance of models trained for classifying the various levels of emotion intensity (task C) and used to predict the binary labels by assigning the positive class to all non-zero intensities, with the performance of models trained directly for the binary classification subtask. We also compare training on either the single-task or the multi-task setting.}
    \label{table:text-only-binary-vs-multi-class}
\end{table}}}

\newcommand{\resultsbymodalities}{
{\renewcommand{\arraystretch}{1.5}
\begin{table}[!ht]
    \centering
    \begin{tabular}{l|cccc}
         \toprule
         Features Used & Task A & Task B & Task C & Mean \\
         \midrule
         Only Text & 0.5127 & 0.6494 & 0.5001 & 0.5541\\
         Only Image & 0.5139 & 0.6404 & \textbf{0.5117} & 0.5553\\
         Only CLIP & 0.5113 & \textbf{0.6559} & 0.4835 & 0.5502\\
         \hline
         Image + Text & 0.5118 & 0.6452 & 0.5041 & 0.5537 \\
         CLIP + Image & 0.5077 & 0.6398 & 0.5053 & 0.5510\\
         CLIP + Text & 0.5118 & 0.6551 & 0.5032 & \textbf{0.5567} \\
         \hline
         \textit{Image + CLIP + Text (submission)} & \textit{\textbf{0.5178}} & \textit{0.6394} & \textit{0.5029} & \textit{0.5534} \\
         \bottomrule
    \end{tabular}
    \caption{Ablation study of our MMMT Transformer. The training conditions are the same except the different input features present. The differences between modalities are very small.}
    \label{tab:modalities-results}
\end{table}}}

\newcommand{\leaderboardtable}{
{\renewcommand{\arraystretch}{1.5}
\begin{table}[!ht]
    \centering
    \begin{tabular}{l|cccc}
         \toprule
         Team Name           & Task A    & Task B    & Task C    & Mean \\
         \hline \hline
         \textit{BLUE (\textbf{our team})}	        & \textit{\textbf{0.5318}}	& \textit{0.8059}	&\textit{ 0.5443}	& \textit{\textbf{0.6273}} \\
         \hline
         BROWALLIA \cite{BROWALLIA}	    & 0.5255	& 0.767	    & 0.5453	& 0.6126 \\
         Amazon PARS \cite{Amazon}	& 0.5025	& 0.7609	& \textbf{0.5564}	& 0.6066 \\
         HCILab \cite{hcilab}	        & 0.4995	& 0.7414	& 0.5301	& 0.5903 \\
         weipengfei	    & 0.4887	& 0.6915	& 0.5033	& 0.5612 \\
         \hline
         BASELINE	    & 0.434	    & 0.7358	& 0.5105	& 0.5601 \\
         \hline
         Yet \cite{Yet}	        & 0.5088	& 0.6106	& 0.51	    & 0.5431 \\
         Greeny	        & 0.5037	& 0.6106	& 0.484	    & 0.5328 \\
         Little Flower \cite{Little}	& 0.5081	& \textbf{0.8229}	& N/A	    & N/A$^*$ \\
         \bottomrule
    \end{tabular}
    \caption{Scores obtained by all of the participating teams, for each task. We also report the average score achieved by the teams over all of the three tasks. Our team obtained the highest average score across the three tasks. $^*$ the team participated only on tasks A and B.}
    \label{tab:leaderboard}
\end{table}
}}

\subsection{Text-Only Experiments}

\textonlymultivssingletable

In order to measure the benefits of multi-task learning for classifying emotion intensities, we performed an ablation study by comparing the weighted F1 scores computed for each emotion intensity predictions made by models trained in two settings. Firstly, we trained independent models for each of the emotion subtasks as defined by tasks A and C. Secondly, we trained a single model in the multi-task setting, by fine-tuning the BERT encoder on all of the emotion subtasks at once.

From Table \ref{table:text-only-single-vs-multi} we can see that, with the exception of sentiment classification, the predictions of all emotion intensities have benefited from being learnt jointly by the model. Following these observations, our submission used the single-task model to predict the sentiment labels for the test dataset (task A), and the multi-task model to predict all other fine-grained emotion labels (task C).

The difference between task B and task C is that for humour, sarcasm and offensive emotions, task B is a binary classification task instead of a multi-class one. The previously defined models trained for task C can be used to make predictions for task B, by mapping the non-zero intensity predictions to the positive class in the binary setting. Because of this, we wanted to see if we could gain any performance improvement by training the classification heads specifically for the classification defined by task B.

Table \ref{table:text-only-binary-vs-multi-class} displays the performance obtained for each setting. When training in the multi-task paradigm, we use the same model definition and training strategy and we only change the classification heads' output sizes to $2$ (for humour, sarcasm and offensive emotions). We observe that the best performing method is training single-task models for each of the three subtasks. Thus, we used these models' predictions on the test dataset for our task B submission.

\textonlybinaryvsmulticlasstable

\subsection{Multi-Modal Experiments}

Table \ref{tab:modalities-results} showcases an ablation study performed on the different modalities. The architecture remains the same in all situations. While the results are close, it is clear that the addition of CLIP image features in the computation positively improved overall results. For the final submission, we used all the available modalities.

\resultsbymodalities

\subsection{Shared Task Results}
\resultsbyemotiontable

We report in Table \ref{table:results-by-emotion} the scores obtained by both of our approaches, separately for each emotion subtask from the test dataset. We observe that none of the models outperforms the other on all subtasks. The text-only approach seems to be better suited for identifying humour and offensiveness, while the multi-modal model performs better on all of the other emotions. Looking at Table \ref{table:results-by-task}, even when comparing the models by their results on the three main tasks, the multi-modal approach does better only on tasks A and C.

We also provide the scores achieved by all participating teams for each task in Table \ref{tab:leaderboard}. Our team managed to place first for task A, second for task B and third for task C. Moreover, our team obtained the highest average score across the three tasks.

\resultsbytasktable

\leaderboardtable

\section{Conclusions \& Future Work}
This work presented team BLUE's approach for the 2022 edition of the MEMOTION 2.0 workshop. We described two solutions for meme classification: i) text-only approach through fine-tuning a BERT model and ii) a Multi-Modal-Multi-Task transformer network that operates on both images and text. Different from most previous methods, we employed CORAL \cite{coral2020} for performing ordinal regression for ordinal outputs (e.g. humour intensity).

By making use of powerful, state-of-the-art, pretrained models for text and images, we obtained the first place on task A with a weighted F1 score of 0.5318, second place on task B with a score of 0.8059 and third place on task C with a score of 0.5453. In addition, we obtain the highest average score for all three tasks.

For future work, we aim to address the issue of severely imbalanced training data and small amount of images by designing a pipeline for self-supervised pretraining on internet-meme images. By only fine-tuning on a small and densely annotated images, the model is more robust to overfitting and predicting the majority class in training.

Moreover, the text in the MEMOTION 2.0 dataset is cleaned by human annotators. However, for a large-scale meme dataset used for pretraining, one can employ lexical normalization models \cite{DBLP:journals/corr/abs-1710-03476,bucur2021sequencetosequence} to automatically correct faulty OCR and transform the text to its canonical form, which was a significant problem in computational pipelines from the first edition of this shared task. 

\bibliography{refs}

\end{document}